\documentclass[conference, letterpaper]{IEEEtran}
\usepackage{times}
\usepackage{helvet}
\usepackage{courier}
\usepackage{tikz}

\def\checkmark{\tikz\fill[scale=0.4](0,.35) -- (.25,0) -- (1,.7) -- (.25,.15) --
cycle;}

\ifCLASSINFOpdf
\else
\fi
\usepackage{subcaption}

\usepackage[cmex10]{amsmath}
\usepackage{color}
\usepackage{fancyhdr}

\renewcommand{\thispagestyle}[2]{}

\fancypagestyle{plain}{
        \fancyhead{}
        \fancyhead[C]{}
        \fancyfoot{}
        \fancyfoot[C]{first page center footer}
}
\begin{document}

%
\title{Phoenix: A Self-Optimizing Chess Engine}

 \author{\IEEEauthorblockN{Rahul Aralikatte}
 \IEEEauthorblockA{IBM India Research Labs\\
 Bangalore, India\\
 Email: rahul.a.r@in.ibm.com}
 \and
 \IEEEauthorblockN{G Srinivasaraghavan}
 \IEEEauthorblockA{International Institute of Information Technology\\
 Bangalore, India\\
 Email: gsr@iiitb.ac.in}}


%


\maketitle

\begin{abstract}
Since the advent of computers, many tasks which required humans to spend a lot
of time and energy have been trivialized by the computers' ability to perform
repetitive tasks extremely quickly. Playing chess is one such task. It was one of 
the first games which was `solved' using AI. With the advent of deep learning, 
chess playing agents can surpass human ability with relative ease. However algorithms 
using deep learning must learn millions of parameters. This work looks at the game of 
chess through the lens of genetic algorithms. We train a genetic player from scratch 
using only a handful of learnable parameters. We use Multi-Niche Crowding to optimize 
Positional Value Tables (PVTs) which are used extensively in chess engines to evaluate 
the goodness of a position. With a very simple setup and after only 1000 generations of
evolution, the player reaches the level of an International Master.
\end{abstract}



%
\IEEEpeerreviewmaketitle

\section{Introduction}
The most important factor that influences the strength of a chess engine is the way in which it
evaluates moves. Expert chess players master the game over years of learning and when they play the
game their mental processes involve a complex synergy of computing possible board moves, memory from
past games, heuristic thumb rules, watching for established patterns of play, etc. Invariably when
humans play the game, the evaluation of a board position or a move is carried out over a complex
hierarchy of abstractions which they are often unable to describe accurately. Trying to replicate or
mimic these processes on a computer is futile given our current lack of understanding of these
processes and abstractions that humans use. Most computer chess engines today resort to a range of
heuristic tricks to somehow reduce the representation of a board position/move to a group of numbers.
Every positional parameter which might influence its goodness must have a numerical equivalent in
this group. The construction of such a group/vector of number is non-trivial. There are 2 main reasons for this:
\begin{itemize}
  \item A small change in the position can result in a large change in its
  goodness. For instance, 2 positions which are identical except for the position of a queen, which
  is offset by a single square, might differ largely in their goodness.
  \item The value of a position might vary from player to player as it depends
  on the goal (s)he sets for himself. That is, a position might be losing for a defensive player and
  might be drawable for an attacking player.
\end{itemize}
Most chess engines today use brute force methods to simulate this thinking
process by taking into consideration multiple parameters which influence the
value of a position. But these parameters are decided by the programmers while
coding and not by the engine itself. Here, an effort has been made to free the
engine from such preconcieved notions and let it learn these parameters and
their relative importance by itself using genetic algorithms.

\section{Evaluation Routine}
At its simplest, an evaluation function of a chess position returns the
\textit{material difference} between the players. However it is generally not
possible to claim equality of two positions taking into account only the
material balance. In several opening lines, one side is ready to sacrifice a
pawn on purpose; for e.g. king's gambit accepted (\textit{1.
e4 e5 2. f4 exf4}). Sometimes a player can sacrifice pieces (like exchanging an
inactive rook for an active bishop or knight) for achieving some non-material
advantage that is considered worthwhile. Moreover, highly skilled chess players
often agree to call the game a draw even when there is material imbalance. So in
all these situations, other factors apart from the material balance have to be
taken into consideration during position evaluation. These factors are known as
\textit{strategic} or \textit{positional parameters}.

In general, the evaluation function is a multivariate, linear function which
measures the goodness of a chess position. There are various
features\footnote{Here the terms feature and parameter are used interchangably}
which can be extracted from a chess position that will give us some insight into
the goodness. The inputs to the evaluation function are numbers which quantify
these features. The output is a single number called the \textit{Evaluation
Score}.

\[F = \sum_{i = 0}^{N} x_i \cdot v_i, \qquad \left\{
  \begin{array}{lr}
    x_i \in \{0, 1\}\\
    v_i \in \mathbf{R}
  \end{array}
  , i = \overline{1, N}
\right.
\]

Here $x_i$ indicates the presence of the $i^{th}$ parameter(feature) and $v_i$
represents its importance as a real number. The position is good for white if
this score is positive and vice versa. Also in some programs, a positive score
is considered to be good for the current player and a negative score for the
opponent.

\subsection{Positional Parameters}
In non trivial chess engines, it is usually the positional parameters which play
an important role in evaluation and give the engine an edge during gameplay.
With the amount of computing resources available today, these parameters when
combined with fast pruning~\cite{fuller73} and deepening~\cite{korf85} techniques
can easily achieve IM\footnote{International Master}  if not the GM\footnote{Grandmaster} level.
Therefore it is important to incorporate them into any new chess program if the
aim is to build a competitive engine. The most important ones are listed
below:
\begin{itemize}
  \item \textbf{Castling}: It is very important for the king to be defended by 
  friendly pieces in the opening and middle games. Pawns in the corner serve as 
  excellent defenders.
  \item \textbf{Rook on an open file}: Rook is the second most powerful piece on
  the board after the queen and has a long reach. But this reach is useless if 
  there are other pieces blocking its way. Therefore rooks should ideally
  be placed on open files which helps a player to initiate an attack.
  \item \textbf{Rook on a semi-open file}: A semi-open file is one in which there are
  only enemy pawns blocking the rook. Placed onto a semi-open file, a rook does
  not allow the opponent to leave his pawn unprotected, thus reducing the
  mobility of opponent's pieces.
  \item \textbf{Knight's mobility}: A knight's mobility is directly dependant on its
placement. A knight near the center of the board is far more valuable than one at the corners.
  \item \textbf{(Supported) knight/bishop outpost}: Outposts are those squares on the
  opponent's side of the board where a piece cannot be attacked immediately.
  These squares act as launch points for a mating attack.
  \item \textbf{Bishop pair}: A bishop pair is generally very useful in end games when
there are few pieces left on the board. They are generally used in tandem to
force the king out of outposts, break pawn chains and hinder the movement of passed pawns.
  \item \textbf{Center pawns (\textit{d4, d5, e4, e5})}: The main theme during openings is \textit{center
control} which results in healthy development of the minor pieces and hinder the
opponent's development. This is largely achieved using the `d' an `e' pawns.
  \item \textbf{Doubled pawns}: A good pawn structure is very important during
  middle and end games. Doubled pawns are usually considered to be a weakness.
  When doubled, the upper pawn blocks the lower and in many cases they need to be defended by a piece.
  \item \textbf{Backward pawn}: A backward pawn generally holds an entire pawn
  chain in place. It is the pawn of a chain which is nearest to the back rank.
  Usually another piece must guard this against attacks which can be disadvantageous.
  \item \textbf{Rook(s) on the $7^{th}$ rank}: Rook(s) on the $7^{th}$ rank
  impede the movement of the opponent's pieces and also act as a very strong
  launch pad for the start of a mating attack as the opponent's king will
  generally be confined to the $8^{th}$ rank.
  \item \textbf{Connected rooks}: When two rooks are on the same file/rank
  without pawns or pieces between them, they are said to be connected. Connected rooks 
  are untouchable as they support each other and can cause huge damage.
  \item \textbf{Passed pawn}: A pawn is said to be passed when there are no enemy pawns
on its file or on either of the adjacent files. This pawn is usually considered very 
important as it has a high probability of getting promoted during end games.
  \item \textbf{Rook-supported passed pawn}: When there is a passed pawn on the board, a
general rule of thumb is to put a rook behind it. This prevents the opponent
from capturing the pawn easily and forces him to block the pawn with a piece.
  \item \textbf{Isolated pawns}: An isolated pawn has no friendly pawns on
  either of the adjacent files. This becomes a weak point in the game as it
  requires constant support from one of the other pieces.
  \item	\textbf{Bishops on the large diagonals}: \textit{a1-h8} and
  \textit{h1-a8} are known as the large diagonals. When bishops are placed along
  these diagonals, they can be tucked into corners while maintaining their
  reach. This is greatly used to mount an attack on the enemy king during
  closed\footnote{A position is said to be closed when there are 6 or more pawns
  occupying the 16 central squares}  middle games.
\end{itemize}

There are many other parameters which can be considered like pawn structures and
closeness of positions. But the ones mentioned above can be easily represented
using Positional Value Tables that will greatly simplify the
formulation and optimization processes which is the main aim of this paper. 

\section{Genetic Optimization}
\subsection{Genetic Scheme}
Genetic Algorithms are mainly used for optimization and their learning is
loosely based on several features of biological evolution. Generally they
require 5 components~\cite{DavidJ}:
\begin{itemize}
  \item A way to encode the solutions of a problem on
  chromosomes\footnote{Chromosomes can be thought of as the fundamental
  building blocks of potential solutions consisting of the solutions'
  parameters.}.
  \item A fitness function which returns a rating for each chromosome.
  \item Operators that are applied on parent solutions when they reproduce to
  alter their genetic composition. Crossover and Mutation are the most common
  operators. Domain-specific operators can also be constructed and used(as
  done in this paper).
  \item Parameter settings for the algorithm like the population size, number of
  generations and so forth.
\end{itemize}

When a genetic algorithm is run using a suitable solution representation the
algorithm can produce populations whose individuals get better and better with
time, finally converging to a solution which is close to the optimum.

But this approach is not sufficient here as there is no one right way to play
chess. The style of a player who plays attacking chess cannot be compared to
that of a player who prefers defensive positional play as both may be equally
good. It is therefore safe to assume that, a mathematical function representing
position evaluation (if it exists) is multimodal.

\subsection{Niching}
The problem with simple genetic algorithms is that it will eventually
converge to one of the many global optima (if they do not get trapped in local
optima) which depend on the initial population and the random genetic
drift~\cite{42} occurring throughout the run. Eventually we will get copies of
the same individual in one of the valleys/plateau.

\begin{figure}[ht]
  \centering
   \includegraphics[scale=0.6]{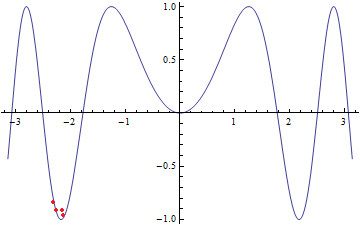}
    \caption{Potential solutions using simple GA}
     \label{fig:sin_simple}
\end{figure}

For instance, consider a simple function \begin{math}f(x)=sin(x^2)\end{math}. A
plot of the function along with the position of the individuals (red marks) trying to find the
minimum is shown in Fig. \ref{fig:sin_simple}.

This function has more than one minimum and it would be ideal if
both the minima were found. This is where the concept of \textit{niching}
comes in handy. Niching is a general class of techniques that promote the formation and
maintenance of stable sub-populations in a genetic algorithm. 2 main objectives
of such techniques are:

\begin{itemize}
  \item To converge to multiple, highly fit, and signiï¬�cantly different
  solutions (for multimodal optimization) 
  \item To slow down convergence in cases where only one solution is required
  (to avoid premature convergence)
\end{itemize}

\begin{figure}[ht]
  \centering
   \includegraphics[scale=0.6]{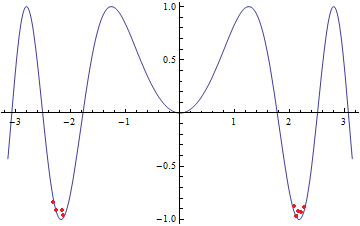}
    \caption{Potential solutions using niching}
     \label{fig:sin_niching}
\end{figure}

Fig. \ref{fig:sin_niching} shows how individuals have converged onto 2 separate
optima for \begin{math}f(x)=sin(x^2)\end{math} when using niching. There are
many approaches to niching and the one used here is called \textit{multi-niche crowding
(MNC)}.

\subsubsection{Multi-Niche Crowding}
Crowding~\cite{44} is a generalization of pre-selection. In crowding, the
selection and reproduction processes are the same as those carried out in simple
genetic algorithms, but the replacement process is different. Assuming that 2
parents produce 2 offspring, in order to make room for the newborns, it is
necessary to identify 2 members from the population for replacement. The policy
of replacing a member of the present generation by an offspring is carried out
as follows:
\begin{itemize}
  \item A group of $C$ individuals is selected at random from the population.
  $C$ is called the crowding factor and a value 2 or 3 appears to work well
 in~\cite{44}.
  \item The chromosomes of the offspring are compared with those of the $C$
  individuals in the group using Hamming distance as a measure of similarity. 
  The group member which is most similar to the offspring is replaced by the offspring.
  \item This procedure is repeated for the other offspring as well. 
\end{itemize}

Crowding is essentially a successive replacement strategy. This strategy
maintains the diversity in the population and postpones premature convergence.
However generic crowding cannot maintain stable subpopulations for long due to
\textit{selection pressure}\footnote{Any cause that reduces reproductive success in a
portion of the population}.

In multi-niche crowding (MNC), both selection and replacement steps are modified
with some type of crowding. The idea is to eliminate the selection pressure
caused by \textit{fitness proportionate reproduction (FPR)} while allowing the population
to maintain some diversity. This objective is achieved in part, by encouraging
mating and replacement within members of the same niche while allowing for some
competition for slots among the niches. The result is an algorithm that (a)
maintains stable subpopulations within different niches, (b) maintains diversity
throughout the search, and (c) converges to different optima.

In MNC, the FPR selection is replaced by what is called \textit{crowding
selection}. In crowding selection, each individual in the population has the
same chance for mating in every generation. Application of this selection rule
takes place in two steps. First, an individual $A$ is selected for mating. This
selection can be either sequential or random. Second, its mate $M$ is selected,
not from the entire population, but from a group of individuals of size $C_s$,
picked at random (with replacement) from the population. The mate $M$ thus
chosen must be the one which is the most `similar' to $A$. The similarity metric
used here is not a \textit{genotypic} metric such as the Hamming distance, but a
suitably defined \textit{phenotypic} distance metric. Crowding selection
promotes mating between individuals from the same niche while allowing mating
between individuals from different niches.

During the replacement step, MNC uses a replacement policy called \textit{worst among
the most similar}. The goal of this step is to pick an individual from the
population for replacement by an offspring. Implementation of this policy
follows these steps. First, $C_f$ groups are created by randomly picking \textit{s}
individuals (with replacement) per group from the population. These groups are
called \textit{crowding factor groups}. Second, one individual from each group that is
most phenotypically similar to the offspring is identified. This gives $C_f$
individuals that are candidates for replacement by virtue of their similarity to
the offspring that will replace them. From this group of most similar
individuals, we pick the one with the lowest fitness to die and that slot is
filled with the offspring. The offspring could possibly have a lower fitness
than the individual being replaced~\cite{45}.

\subsection{Problem Formulation}
Usually evaluation functions consider almost if not all the parameters discussed
in the Positional Parameters section. But here, the evaluation has to be totally
dependent on whatever the computer learns on its own and nothing else. Therefore
the formulation of the problem had to be done in such a way that the engine
itself recognizes the relative importance of each parameter. This requirement
forced the use of PVTs\footnote{Positional Value Tables}.

\begin{figure}[ht]
  \centering
   \includegraphics[scale=0.8]{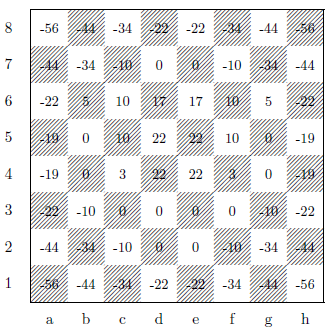}
    \caption{PVT for a Black Knight}
     \label{fig:pvt_knight}
\end{figure}

A PVT is a grid of numbers which indicate the best squares for a piece to
occupy. Greater the number, better is its position. For example, Fig.
\ref{fig:pvt_knight} shows the PVT for a black knight. It is easy to spot that
the values at the center of the board are higher than those at the corners. This
means it is desirable for the knight to be on the central squares rather than
the corners.

\begin{table}
	\centering
	\label{tab:PVTs}
    \begin{tabular}{| c | c | c |} \hline
    \textbf{Piece} & \textbf{Middle Game} & \textbf{End Game} \\ \hline
	  Pawn & \checkmark & \checkmark \\ \hline
	  Rook & \checkmark & $\times$ \\ \hline
	  Knight & \checkmark & \checkmark \\ \hline
	  Bishop & \checkmark & \checkmark \\ \hline
	  Queen & \checkmark & $\times$ \\ \hline
	  King & \checkmark & \checkmark \\ \hline
    \end{tabular}
    \caption{PVTs}
\end{table}

By carefully constructing these PVTs, most of the positional
parameters can be quantified. Even though some of the abstract
ones such as rook behind a past pawn cannot be described using this technique
alone, by combining this with other search algorithms, it is possible to get
more accurate evaluation scores. This is achieved by using PVTs for move
ordering~\cite{newborn77} by arranging moves in descending order of the PVT values for
the destination squares.

It is easy to see that each piece must have its own PVT. Also, most of the
pieces have different responsibilities at different junctures of the game.
Therefore it is necessary to maintain different PVTs for middle and end games
(PVTs for openings are not required as moves are made using opening books).
Table 1 shows the various PVTs that are maintained.

\begin{figure}[ht]
  \centering
   \centerline{\includegraphics[scale=0.5]{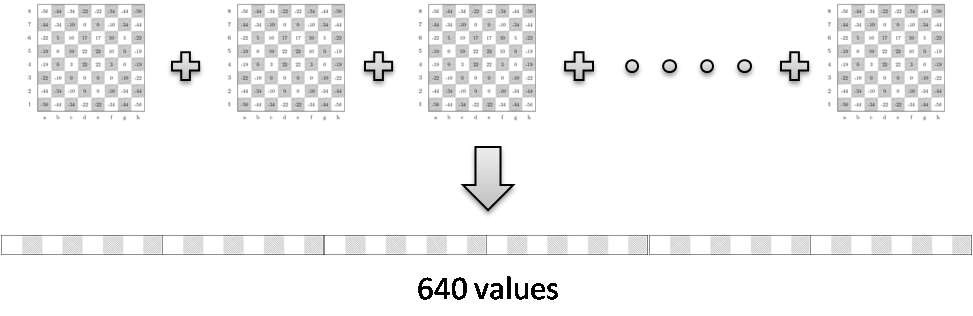}}
    \caption{PVTs to Chromosome}
     \label{fig:2d_to_1d}
\end{figure}

Every potential solution (player/individual in the population) should have a
chromosome which will influence its gameplay. A good chromosome will often lead
to victory where as a bad chromosome will result in a loss. Therefore each
value in every PVT is considered as a parameter to be optimized and hence
all the PVTs are joined together to form a chromosome as shown in Fig.
\ref{fig:2d_to_1d}.

This is essentially a single dimension array containing 640 floating point
numbers (clubbing together the 10 PVTs mentioned in Table 1, each having 64
values). From here on, the problem of learning reduces to an optimization of
these 640 values.

\subsection{Implementation}
CuckooChess~\cite{40} is an advanced free open source chess program under the
GNU General Public License written in Java by Peter Osterlund.  It contains many
of the standard algorithms for computer chess discussed previously such as
iterative deepening, quiescence search with SEE pruning, MVV/LVA move ordering,
hash table, history heuristic, recursive null moves, opening book and magic bit
boards. It also uses some advanced techniques like Negascout, aspiration
windows, futility pruning and late move reductions~\cite{41}. Version 1.12 of this engine has
been modified suitably to achieve the intended results. These modifications
include:
\begin{itemize}
  \item Replacing the entire evaluation module with a new module which utilizes the PVTs learnt by the system
  \item Adding a genetic training system with a tournament selector which is used to learn the PVTs
  \item Integrating the learning and playing pipelines
  \item Modifying search behavior to take advantage of the new PVT based evaluation routine
  \item A tournament simulator module which is used for both fitness evaluation and testing
\end{itemize}

\begin{figure}[ht]
  \centering
   \centerline{\includegraphics[scale=0.5]{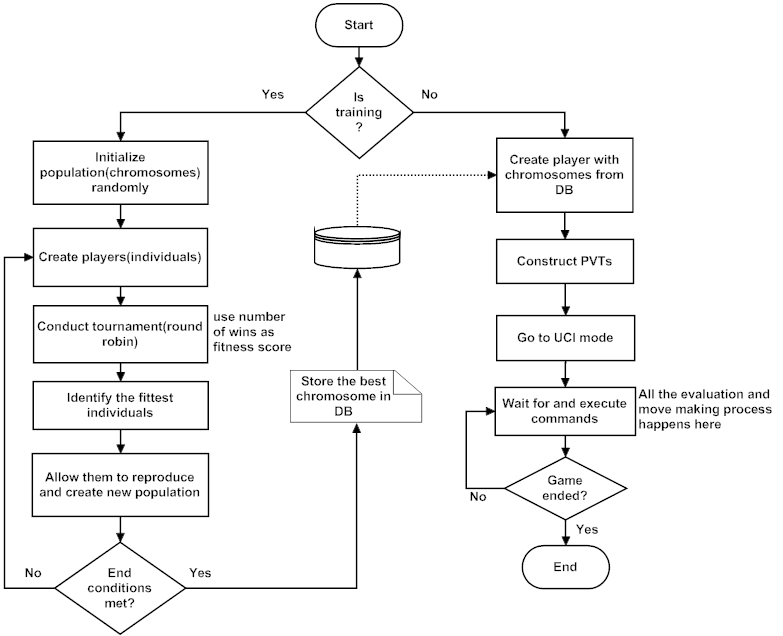}}
    \caption{Workflow}
     \label{fig:phoenix_flowchart}
\end{figure}

Fig. \ref{fig:phoenix_flowchart} is a flowchart which gives an overview of what
 is going on under the hood.

First, a population of 20 players is created. The players' chromosomes are
randomly initialized. At this stage, the players make random and losing moves.
They are pit against each other in a round-robin tournament where each player
plays at least 3 games and the results are recorded. The players get 1 point for
a win, 0.5 for a draw and 0 for a loss. After the tournament is finished, the
results are tabulated and these points are used as an indicator of their
chromosomes' fitness level.

MNC is used to optimize the compressed PVTs. The individuals start converging
towards optimal solutions which are stored in a database of chromosomes once any
of the following end conditions are me:
\begin{itemize}
  \item 1000 generations have passed
  \item The best solution remains the same for 10 or more generations
  \item The rate of change in the best solution chromosome structure is below
  1\% for 20 or more generations
\end{itemize}
When the chess engine is asked to play, it selects one from the database and
re-forms PVTs from the chromosome and uses it to make moves.

\section{Results}
The best way to test a chess engine is to make it play against other engines.
This can be daunting and painfully slow if there is no common language through
which the engines can communicate. In chess programming, there exist 2 protocols
which are used as standards while building chess engines. They are:
\begin{itemize}
  \item Universal Chess Interface (UCI)
  \item Chess Engine Communication Protocol (used in XBoard and WinBoard)
\end{itemize}
UCI is more robust and is supported by most of the prominent engines today.
Therefore UCI is used as the communication protocol in this implementation. More
information about this protocol can be found in~\cite{46}.  Many chess GUIs also are UCI compatible and
hence it is easy to plug the engine into a GUI such as Arena~\cite{47} and
actually see the games being played, rather than read the PGN.

The engine was tested against the original CuckooChess engine as it would
provide a clear benchmark about any improvements achieved. A total of 1000 games
were played and recorded with a time control of 3 seconds per move.
The games were then analyzed using the EloStat algorithm. This algorithm
calculates the Elo rating~\cite{48} of a player provided the rating of the
opponent is known. The results are tabulated below.

\begin{figure}[ht]
  \centering
   \centerline{\includegraphics[scale=0.25]{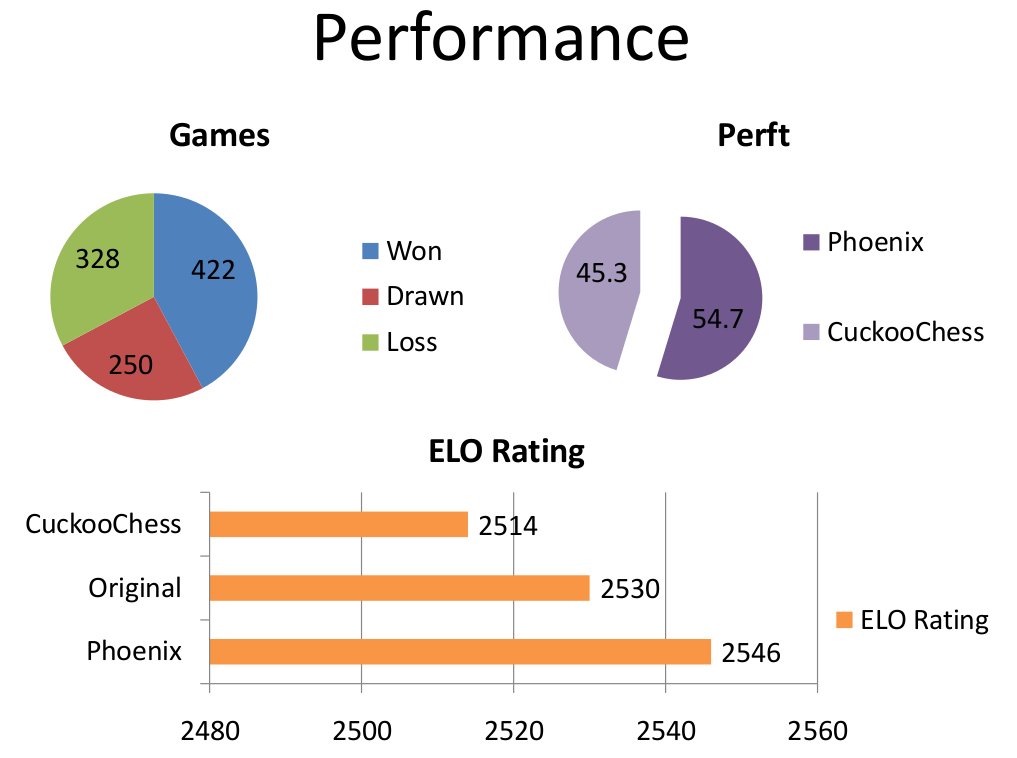}}
    \caption{EloStat Result}
     \label{fig:result}
\end{figure}

CuckooChess is rated at 2530 according to CCRL (Computer Chess Ratings List). 
After 1000 games, it was seen that the modified engine outperformed its parent 
with a rating of 2546 (an increase of 19 points). This rating puts Phoenix 
in the `International Grandmaster' category.

Even though the increase in rating seems small, it should be noted that the rise
in rating is tapered off gradually when large number of games are played
with an opponent of similar strength. It can be seen that Phoenix has won
422 games compared to the 328 games won by CuckooChess. This indicates that
there is a considerable increase in strength of the modified engine.

Also it has to be noted that the solution we obtained is not optimal. With
logging of the obtained chromosomes, it can be seen that some parts of the PVTs
seem random. This is because it is not possible for the algorithm to find out
which mutations will help in pushing the individual in the correct direction and
which do not. This results in the replacement of some good values with
random ones. Therefore there is still room for improvement in the engine if 
mutation techniques can be enhanced.




\bibliographystyle{IEEEtran}
\bibliography{IEEEabrv,ref}
%

\end{document}